\documentclass{article}

\usepackage[preprint]{neurips_2024}

\usepackage[utf8]{inputenc} 
\usepackage[T1]{fontenc}    
\usepackage{hyperref}       
\usepackage{url}            
\usepackage{booktabs}       
\usepackage{amsfonts}       
\usepackage{nicefrac}       
\usepackage{microtype}      
\usepackage{xcolor}         
\usepackage{accvabbrv}

\usepackage{graphicx}
\usepackage{multirow}

\usepackage[accsupp]{axessibility} 

\usepackage{orcidlink}

\title{DeTurb: Atmospheric Turbulence Mitigation with Deformable 3D Convolutions and 3D Swin Transformers}

\author{%
  Zhicheng Zou and Nantheera Anantrasirichai\thanks{This work was supported by the UKRI MyWorld Strength in Places Programme (SIPF00006/1).} \\
  Visual Information Laboratory\\
  University of Bristol\\
}

\begin{document}

\maketitle

\begin{abstract}
Atmospheric turbulence in long-range imaging significantly degrades the quality and fidelity of captured scenes due to random variations in both spatial and temporal dimensions. These distortions present a formidable challenge across various applications, from surveillance to astronomy, necessitating robust mitigation strategies. While model-based approaches achieve good results, they are very slow. Deep learning approaches show promise in image and video restoration but have struggled to address these spatiotemporal variant distortions effectively. This paper proposes a new framework that combines geometric restoration with an enhancement module. Random perturbations and geometric distortion are removed using a pyramid architecture with deformable 3D convolutions, resulting in aligned frames. These frames are then used to reconstruct a sharp, clear image via a multi-scale architecture of 3D Swin Transformers. The proposed framework demonstrates superior performance over the state of the art for both synthetic and real atmospheric turbulence effects, with reasonable speed and model size.
\end{abstract}

\section{Introduction}
\label{sec:intro}

Light propagation through the various layers of the atmosphere, which differ in temperature, pressure, humidity, and wind speed, introduces diffraction-related blurring and random refractions. These variations cause fluctuation in intensity and random phase distortions in the wavefront of light waves, significantly degrading the performance of imaging systems. Most image and video enhancement and restoration techniques have been proposed, addressing specific problems like denoising and deblurring. However, these methods may not be directly used to solve the problem of atmospheric turbulence, as it involves multiple types of distortion, making it challenging to model its degradation accurately. Successful approaches will be invaluable in many applications, including air-to-ground imaging, long-range terrestrial video surveillance, creative industries such as natural history filmmaking, and other computer vision applications, including object recognition and tracking.

Traditional methods for turbulence video restoration typically involve: (i) removing pixel offset caused by tilt, often using the optical flow method; (ii) employing lucky image fusion to select and combine the clearest pixel blocks within a specific interval; and/or (iii) applying blind deconvolution algorithms to remove residual fuzziness~(\cite{photonics10060666, 10.1117/12.896183, 6976872, 6178259, CHEN2020106131}). However, these methods have notable limitations. They rely heavily on a large quantity of measured data for fusion, use general and non-optimized point spread function (PSF) priors in blind deconvolution, and struggle to effectively address the complex statistical behavior of atmospheric turbulence~(\cite{photonics10060666, 10.1117/1.OE.57.2.024101, Fried:78}). Additionally, they are prone to artifacts from inaccurate flow estimates and are very slow.

In recent years, deep learning approaches have emerged to tackle the challenge of atmospheric turbulence mitigation. These methods include deep-stacked autoencoder neural network models and convolutional UNet-like architectures~(\cite{gao2019atmospheric,photonics10060666,ANANTRASIRICHAI202369}). Although these models have shown promising results in simulations, they often rely on simplified assumptions about atmospheric turbulence and lack sufficient real-world datasets, limiting their generalization to diverse scene reconstructions. The effectiveness of these models on actual measured data remains an area of active research.

In this work, we propose a novel framework called DeTurb that integrates two modules: (i) a \textit{non-rigid registration} module to reduce wavy effects and temporal distortion caused by atmospheric turbulence, and (ii) a \textit{feature fusion} module to select and fuse useful features for enhanced visualization. Several existing video restoration methods include alignment and feature fusion modules~(\cite{Wang:EDVR:2019, Jiang_2023_CVPR, lin2024bvi, Lin:Spatio:2024}), which we found to be more effective when separated rather than combined into a single process. The two-step approach has also proven effective in dealing with atmospheric turbulence issues. Model-based methods, such as space-invariant deconvolution (SID) (\cite{6178259}) and complex wavelet-based fusion (CLEAR) (\cite{Anantrasirichai:Atmospheric:2013}), employ non-rigid image registration to reduce random perturbations. SID then applies deblurring to the near-diffraction-limited image, while CLEAR produces a sharp image through wavelet-based image fusion. With the advent of deep learning, similar strategies have continued (sometimes referred to as tilt-blur models), such as the CNN-based method AT-Net (\cite{Yasarla:ATNet:2021}) and the transformer-based method TMT (\cite{Zhang_TMT}).

In our DeTurb framework, both modules employ UNet-like architectures. The first module learns different levels of turbulence distortion, while the second module extracts features from different scales, reconstructing local details related to semantic meanings.  A multi-scale approach (like UNet) is essential because the levels of turbulence distortion are unpredictable (atmospheric turbulence is quasi-periodic \cite{Li:Suppressing:2009}). Additionally, distortion levels can vary spatially due to the varying distances between objects and the camera—the further an object is to the camera, the more distortion it shows. 
To address displacement among frames caused by turbulence and moving objects, deformable convolutions were applied in~(\cite{hu:object:2023}). However, our architecture differs by processing data in a 3D manner, handling features in both spatial and temporal dimensions simultaneously. Fluctuations in velocity due to atmospheric turbulence cause local displacements between frames, rather than uniform global shifts~(\cite{Huebner:compensating:2009}). As a result, 3D operations are more effective than 2D ones in extracting and enhancing features. Specifically, non-rigid registration is achieved using deformable 3D convolutions~(\cite{9153920}). After local spatial alignment among frames, the aligned features from each layer of the pyramid are enhanced using 3D Swin Transformers~(\cite{yang2023swin3d}).

In summary, our main contributions can be summarised as follows:  
\begin{itemize}
    \item We propose a novel framework, DeTurb, for restoring long-range videos affected by spatiotemporal distortions due to atmospheric turbulence. With comparable inference speed, DeTurb significantly outperforms the state of the art in terms of video quality.
    \item DeTurb mitigates geometric distortion using a non-rigid registration module, and then enhances edges and texture details with a feature fusion module.
    \item The non-rigid registration module estimates the flow of random perturbations and moving objects via a UNet-like architecture, in which each scale performs deformable 3D convolutions.
    \item The feature fusion module combines features of registered frames with 3D Swin transformers arranged in a UNet-like architecture. This aims to enhance both local and global details for better visualization.
\end{itemize}

\section{Existing methods}
Atmospheric turbulence, usually resulting from temporal variations occurring near the ground, is typically anisoplanatic for large field-of-view objects, exhibiting spatial variations that complicate the correction process. Given these complexities, learning-based techniques, specifically those involving deep learning, have become increasingly effective in mitigating these effects. Unlike traditional methods (e.g.,~(\cite{Anantrasirichai:Atmospheric:2013, 6178259, Kelmelis:practical:2017, Boehrer:Turbulence:2021}), these techniques rely on the processing power of neural networks to predict and correct distortions, leading to more adaptable and robust solutions.

Convolutional Neural Networks (CNNs) have been at the forefront of this effort, as they are well-suited for image processing tasks due to their ability to extract hierarchical features from images. CNNs can effectively learn to recognize and mitigate the effects of turbulence, thereby significantly enhancing image clarity~(\cite{gao2019atmospheric, Li:18, 9762752}). Another innovative approach is Generative Adversarial Networks (GANs), which employ a dual-network architecture comprising a generator and a discriminator that work in tandem to produce highly refined outputs from severely degraded inputs. GANs are capable of generating clear, high-resolution images from those distorted by atmospheric conditions. For example, ATVR-GAN (\cite{s23218815}) integrates a Recurrent Neural Network (RNN) into the GAN's generator, while in (\cite{leihong_zhixiang_hualong_zhaorui_kaimin_dawei_2021}), phase disturbance reduction is performed in the Fourier domain, which is then used as a condition for the GAN. LTT-GAN (\cite{10023498}) applies style transfer via GAN to restore faces degraded by atmospheric turbulence.

More recently, transformer-based methods have shown promising results in handling atmospheric turbulence by employing different attention mechanisms to process image patches. This enables precise correction of atmospheric distortions across various scales within an image. These methods have discovered long-term dependencies in data and demonstrated great potential in this field~(\cite{10.1007/978-3-031-19800-7_25, 10.1063/5.0160755, Zhang_TMT, Zhang:23}). TurbNet~(\cite{10.1007/978-3-031-19800-7_25}) extracts features through a transformer UNet-like architecture and utilizes physics-inspired downstream methods to reconstruct clean images. The Swin transformer is employed for estimating turbulent flow in (\cite{10.1063/5.0160755}). TMT (\cite{Zhang_TMT}) utilizes vision transformers to remove blur, while ASF-Transformer (\cite{Zhang:23}) integrates spatial-aware and frequency-aware transformer blocks into a UNet framework.

Diffusion models (DMs) are also of interest in these applications. Two recent methods, (\cite{Nair_2023_WACV}) and (\cite{Suin_2024_WACV}), focus on restoring faces degraded by atmospheric turbulence using the Denoising Diffusion Probabilistic Model (DDPM), initially proposed by \cite{Ho:Denoising:2020}. In (\cite{Jaiswal_2023_ICCV}), a physics-based simulator is directly integrated into the training process of a restoration model. Furthermore, \cite{Wang:Atmospheric:2023} proposes a conditional diffusion model under a variational inference framework for more generic images. DMs serve as generative priors for blind restoration in (\cite{Chung:Parallel:2023}), which exploits a degradation model involving tilt and blur, akin to the concept in TMT (\cite{Zhang_TMT}) that utilizes two concatenated modules for tilt removal and deblurring.

\section{Proposed method}

\begin{figure}[tb]
\centering
	\includegraphics[width=\linewidth]{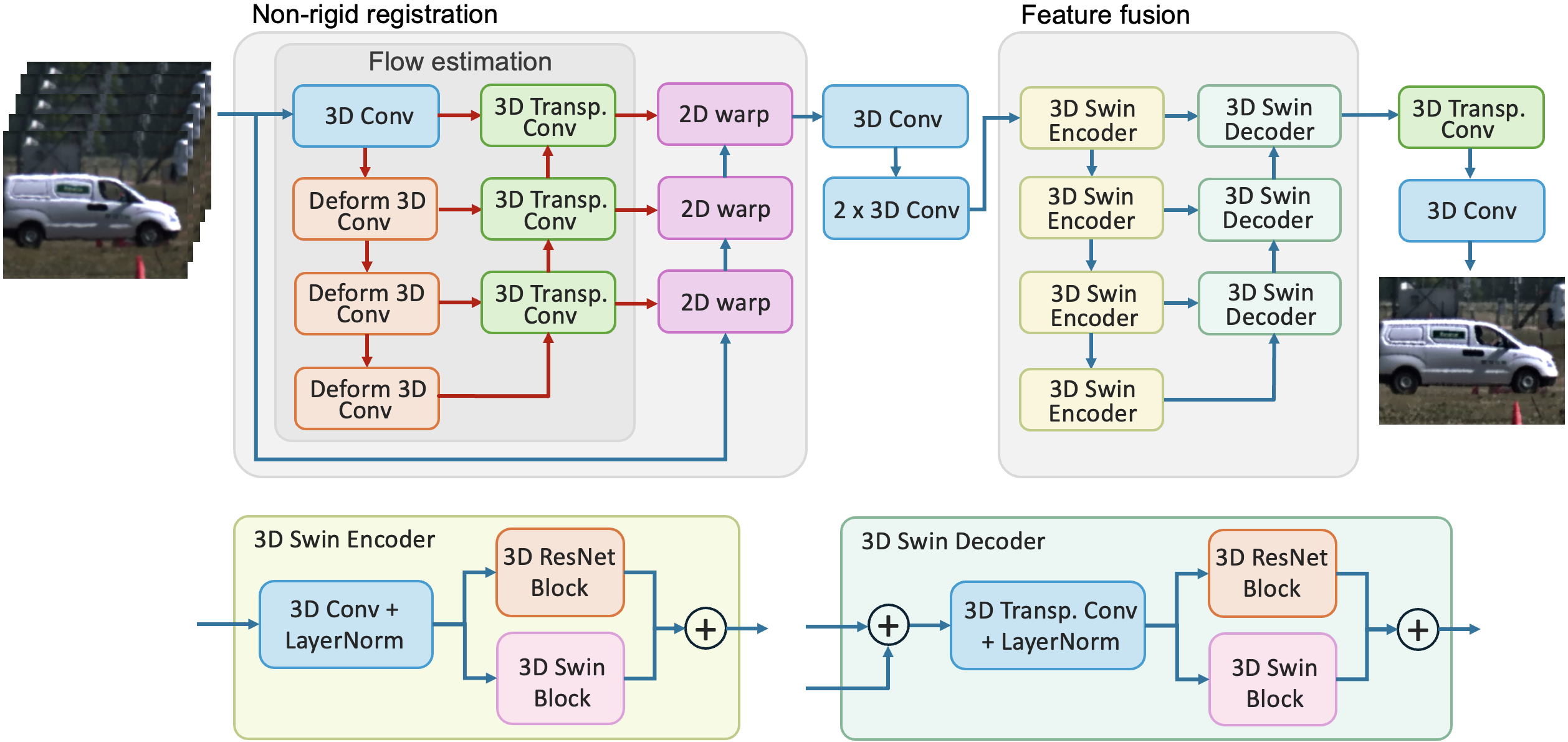}
	\caption{Diagram of the proposed DeTurb. Top-row: end-to-end framework comprising the non-rigid registration and the feature fusion modules. Bottom-row: block diagrams of 3D Swin transformer encoder and decoder blocks.}
	\label{fig:architecture}
\end{figure}

The proposed framework is shown in Fig. \ref{fig:architecture}, where the non-rigid registration module (described in Section \ref{ssec:align}) is concatenated with the feature fusion module (described in Section \ref{ssec:enhance}).

\subsection{Non-rigid registration module}
\label{ssec:align}

As mentioned earlier, several methods address multiple frame inputs of the video using frame or feature alignment processes. Among these, we were inspired by (\cite{Zhang_TMT}), where a UNet architecture with depth-wise 3D convolutions is used to estimate flow across multiple frames. We improve this process with deformable 3D convolutions. A key contribution is that, in atmospheric turbulent environments, objects exhibit visual distortions within small ranges of pixel displacement, appearing randomly in all directions. The use of deformable 3D convolutions provides flexibility in capturing the shapes of the distorted objects across different scales of the UNet in space and time, enabling the extraction of appropriate features from the distorted scenes. After obtaining multi-scale flows each frame in the input group is warped to the current frame from coarse to fine displacements, as shown in Fig. \ref{fig:architecture}.

\paragraph{Deformable 3D convolutions.} The transition from the deformable 2D convolutions (\cite{Dai_2017_ICCV}) to 3D ones involves extending the adaptability of convolution operations to the third dimension. Deformable 3D convolutions are defined as Eq. \ref{DConv3D}, 

\begin{equation}
	y(p_0) = \sum_{p_n \in \mathcal{G}} w(p_n) \cdot x(p_0 + p_n + \Delta p_n),
	\label{DConv3D}
\end{equation}

\noindent where $p_0$ denotes a location in the output feature map $y$, $p_n$ represents the $n$-th location in the convolution sampling grid $\mathcal{G} \in \{(-1, -1, -1), (-1, -1, 0), ... , (1, 1, \\ 0), (1, 1, 1)\}$ for a grid size of 3$\times$3$\times$3, $w$ is the convolution weight, and $x$ is the input feature map. $\Delta p_n$ is the learnable offset for the $n$-th location, introducing adaptability to the convolution operation~(\cite{9153920}). Consequently, the convolutional kernel's receptive field can adjust in response to alterations in the input feature map's shape, thereby accommodating changes in the dimensions and scales of the identified patterns.

\paragraph{Architecture settings.} The basic architecture is a 3D-UNet with a depth of 4. The number of depths of the non-rigid registration relies on the local displacement among frames due to atmospheric turbulence. The deformable 3D convolutions (Def3DConv) are used in depth 2$^\text{nd}$-4$^\text{th}$ as listed in Table \ref{tab: Tilt-Removal} (left). Each convolution is integrated with ReLU activation functions. Deformable convolutions still rely on a predefined kernel size, but it adds the ability to dynamically adjust the sampling locations within that predefined kernel grid during the convolution process. This adjustment allows them to adapt better to local variations in the input data. Following (\cite{Zhang_TMT}) to further enhance the perceptual field and improve the detection of features across varying scales, the kernel sizes are strategically set in the encoder. The first level uses a larger kernel size of 7, maintained for the second level to capture broader features and anomalies caused by turbulence before reducing to size 5 and 3 in the third and fourth levels, respectively. This arrangement allows the network to capture and process various distortion patterns, from broad to more localised disturbances. For decoder, we use 3D transposed convolutions (3DTranspConv) and the kernel size of 3 for all levels.

\begin{table}[tb]
\begin{center}
\caption{Configuration of proposed DeTurb network with the input size of $H\times W \times N$. 3DConv is a 3D convolution block, Def3DConv is a deformable 3D convolution block, and 3DTranspConv is a transposed convolution block. 3DSwinBlock\_Enc is a 3D Swin Transformer Encoder block and 3DSwinBlock\_Dec is a  3D Swin Transformer Decoder block. } 
\resizebox{\columnwidth}{!}{\begin{tabular}{ll|ll}
\toprule \noalign{\smallskip}
\multicolumn{2}{c|}{Non-rigid registration module} & \multicolumn{2}{c}{Feature fusion module} \\
\hline
Layer (kernel size) & Out dimension & Layer (kernel size) & Out dimension \\ \midrule
3DConv+MaxPool (7$\times$7) & $\frac{H}{2}\times\frac{W}{2}\times$64 & 2$\times$3DConv (4$\times$4) & $\frac{H}{2}\times\frac{W}{2}\times$32 \\ 
Def3DConv+MaxPool (7$\times$7) & $\frac{H}{4}\times\frac{W}{4}\times256$ & 3DSwinBlock\_Enc & $\frac{H}{4}\times\frac{W}{4}\times$64 \\
Def3DConv+MaxPool (5$\times$5) & $\frac{H}{8}\times\frac{W}{8}\times256$ & 3DSwinBlock\_Enc & $\frac{H}{8}\times\frac{W}{8}\times128$ \\
Def3DConv (3$\times$3) & $\frac{H}{8}\times\frac{W}{8}\times512$ & 3DSwinBlock\_Enc & $\frac{H}{16}\times\frac{W}{16}\times256$ \\
3DTranspConv+3DConv (3$\times$3) & $\frac{H}{4}\times\frac{W}{4}\times(256 \rightarrow 2N)$  & 3DSwinBlock\_Enc & $\frac{H}{32}\times\frac{W}{32}\times512$\\
3DTranspConv+3DConv (3$\times$3) & $\frac{H}{2}\times\frac{W}{2}\times(128 \rightarrow 2N)$ & 3DSwinBlock\_Dec & $\frac{H}{16}\times\frac{W}{16}\times256$\\
3DTranspConv+3DConv (3$\times$3) & $H\times W\times (64 \rightarrow 2N)$ & 3DSwinBlock\_Dec & $\frac{H}{8}\times\frac{W}{8}\times128$\\
2DWarp$_k$, $k \in \{0,1,2\}$ & $\frac{H}{2^k}\times\frac{W}{2^k}\times 3N$ & 3DSwinBlock\_Dec & $\frac{H}{4}\times\frac{W}{4}\times64$ \\ 
 3DConv (7$\times$7) & $H\times W\times 3N$ & 3DTranspConv & $H\times W\times32$\\
 &  & 3DConv (1$\times$1) & $H\times W \times 3$ \\ 
\bottomrule
\end{tabular}} 
\label{tab: Tilt-Removal}
\end{center}
\end{table}

\subsection{Feature fusion module }
\label{ssec:enhance}

Similar to many methods proposed for video processing (\cite{ANANTRASIRICHAI202369, Shang_2023_CVPR, Lin:Spatio:2024}), we process data through multiscale feature learning using a UNet-like architecture, as shown in Fig. \ref{fig:architecture} and parameters are listed in Table \ref{tab: Tilt-Removal} (right). 
The process begins with input initial feature extraction through two blocks of 3D convolutions, which prepare the data by highlighting essential features for subsequent layers. The 3D Swin Transformer (\cite{yang2023swin3d}) is used as its efficiency in modelling complex dependencies. It utilises shifted window mechanisms to handle the input data's non-uniformity. The processed data is then channelled through a UNet-like architecture with a depth of 4. This approach ensures that the enhancement of local areas is related to the semantic information of those areas and their surroundings.  This multi-scale processing is crucial for restoring fine textures and edges in distorted frames, enhancing the model’s capability to effectively address a range of distortion scales introduced by atmospheric conditions. The last 3D convolution layer converts features to RGB output.

\paragraph{3D Swin Transformer.} The 3D Swin transformers~(\cite{Tang_2022_CVPR, yang2023swin3d, Cai2023SwinUnet3D}) extend the Swin transformer architecture to three dimensions, adapting it to better understand volumetric or sequential data by incorporating the temporal dimension. This offers significant advancements over traditional models in handling the complexities of video reconstruction. The model introduces a hierarchical structure to process data at multiple scales, capturing detailed spatial-temporal features. Our 3D Swin transformer uses a shifted window-based self-attention mechanism across three dimensions as used in (\cite{yang2023swin3d,Cai2023SwinUnet3D}), effectively capturing dynamic changes over time. 

As shown in Fig. \ref{fig:architecture} bottom-row, the 3D Swin Transformer Encoder block uses 3D convolutions to merge and downsample features across multiple channels. These features are then split to undergo processing by one 3D ResNet block and one 3D Swin transformer block, with the final output being the sum of these two blocks' outputs. In each 3D Swin Transformer Decoder block, a 3D transposed convolution operator is used to combine and upsample feature maps. 

Another benefit of the 3D Swin Transformer blocks is that in both the encoder and decoder, cyclic shifts are applied to the input of multi-head self-attention to improve interaction between adjacent and non-adjacent tokens in three directions. It works by shifting the input tokens cyclically by a specified number of units $s$ in each dimension, where $s$ is typically set to half the window size. This shifting rearranges the tokens so that those from adjacent windows in the original configuration may end up in the same window post-shift. This process enables the model to compute similarities and interactions between tokens that were initially in neighboring windows, which helps overcome the limitation of the original window-based self-attention that only computes interactions within the same window. The cyclic shifting thus facilitates the capture of broader contextual information across adjacent windows, enhancing the network's ability to learn long-distance dependency information. However, it also leads to an increase in the number of windows and variability in window sizes, which is managed through a window-masking mechanism to ensure that only relevant token similarities are considered. More details about the 3D Swin transformer block can be found in (\cite{Cai2023SwinUnet3D}).

\subsection{Loss Functions}

Two loss functions are employed similar to (\cite{Zhang_TMT}): Charbonnier Loss and Edge Loss. In the distortion mitigating module, atmospheric turbulence effects can create outliers in the pixel-wise loss. To address this, Charbonnier loss is used, as it combines the benefits of both $\ell_1$ and $\ell_2$ losses, effectively handling outliers better~(\cite{551699}). Defined by Eq. \ref{Loss},

\begin{equation}
	L_\text{Char}(x, y) = \sqrt{(x-y)^2 + \epsilon^2},
	\label{Loss}
\end{equation}

\noindent where $x$ and $y$ represent the predicted and true values, and $\epsilon$ is a small constant (e.g., $1e-3$) to ensure numerical stability, this loss function effectively balances the error distribution. It provides a smooth gradient even when small errors are present, which is particularly beneficial for handling the subtle but critical differences in turbulence-affected images, where precision in error correction is essential. To ensure sharp results, the loss $L_\text{Edge}$ is added after training for a certain number of iterations (300k in this paper). The edge loss $L_\text{Edge}$ is defined as described in Eq. \ref{EdgeLoss},

\begin{equation}
	L_\text{Edge}(x, y) = \lambda L_\text{Char}(x - g((g(x)^{\downarrow2})^{\uparrow2}), y - g((g(y)^{\downarrow2})^{\uparrow2})),
	\label{EdgeLoss}
\end{equation}

\noindent where $g$ is a Gaussian filter, $\lambda$ represents a small gain (set to 0.05 in this paper), and $(\cdot)^{\downarrow2}$ and $(\cdot)^{\uparrow2}$ denote downsampling and upsampling by 2, respectively

\section{Datasets}

\subsection{Synthetic dataset}

As ground truth for atmospheric turbulence mitigation is not available, we generate synthetic distorted sequences from clean ones. We employed the method based on P2S transform, proposed by \cite{Mao_2021_ICCV}, to generate atmospheric turbulence distortions, as it has proved to be efficient for use as training data.

For static scenes, the dataset is sourced from the Place Dataset~(\cite{7968387}), where 9,017 images were randomly selected. These images serve as the basis for simulation, where each generates 50 corresponding turbulence-impacted images and their distortion-free counterparts, resulting in a total of 9,017 pairs of static scene sequences (total 450,850 image pairs). 

For dynamic scenes, the dataset is enriched with video content from multiple sources to ensure variability and complexity, mimicking real-world conditions more accurately. The primary sources for these dynamic scenes are the Sports Video in the Wild (SVW)~(\cite{7163105}) dataset, the ground truth videos from the TSRWGAN~(\cite{jin_chen_lu_chen_wang_liu_guo_bai_2021}) project, and the Video Dataset of Perceived Visual Enhancements (VDPVE)~(\cite{Gao_2023_CVPR}). This integration forms a comprehensive collection of 6,495 video pairs (total 2,749,582 image pairs).

The dataset is split into training and testing subsets to facilitate practical training and evaluation. The static images are divided into 7,499 pairs for training and 1,518 pairs for validation. The dynamic videos are similarly split, allocating 4,700 videos for training and 1,795 for testing, maintaining a cap of 120 frames per video in the testing set to ensure uniformity in evaluation conditions. 

\subsection{Real datasets}

Videos with real atmospheric turbulence are used to evaluate the performance and generalization capabilities of the proposed method. This includes two datasets: the OTIS~(\cite{GILLES201738}) dataset and the CLEAR~(\cite{ANANTRASIRICHAI202369}) dataset. The OTIS dataset contains 16 static scenes with ground truth. The CLEAR dataset includes 11 dynamic scenes with significant motion and 8 static scenes with minimal or no motion. This dataset comes with \textit{pseudo} ground truth generated using complex wavelet-based image fusion. Collectively, these datasets encompass a broad spectrum of turbulence conditions.

\section{Results and discussions}

\subsection{Experiment settings}
We trained the two modules separately. Initially, the non-rigid registration module was trained for 400K iterations using 12 frames of input video, a patch size of 128, and a batch size of 2. This phase focused on establishing a robust base for angular correction before proceeding to more complex tasks. Once this module’s training was solidified, the feature fusion module was trained for 600K iterations under the same batch size, patch size, and learning rate conditions, ensuring consistency in training dynamics. Both models employ the Adam optimizer, known for its efficiency in handling sparse gradients on noisy problems, in conjunction with a Cosine Annealing scheduler. This scheduler adjusts the learning rate from an initial value of $2 \times 10^{-4}$ to $1 \times 10^{-6}$, promoting a gradual and controlled optimization process.

We evaluated and compared our method with four state-of-the-art models designed to mitigate atmospheric turbulence in long-range imaging: AT-Net~(\cite{Yasarla:ATNet:2021}), TurbNet~(\cite{10.1007/978-3-031-19800-7_25}), TSRWGAN~(\cite{jin_chen_lu_chen_wang_liu_guo_bai_2021}), TMT~(\cite{Zhang_TMT}), and DATUM~(\cite{zhang2024spatio}). Additionally, we included the state-of-the-art video restoration models, BasicVSR++~(\cite{Chan_2022_CVPR}), STA-SUNet~(\cite{Lin:Spatio:2024}), and VRT~(\cite{Liang:VRT:2024}). These models were retrained with our synthetic datasets and tested on both synthetic and real-world datasets. All comparative models were trained using similar data augmentation strategies as employed for the TMT, ensuring comparable conditions and fair performance evaluation.

\subsection{Performance on synthetic datasets}

\begin{figure}[tb]
\centering
	\includegraphics[width=\linewidth]{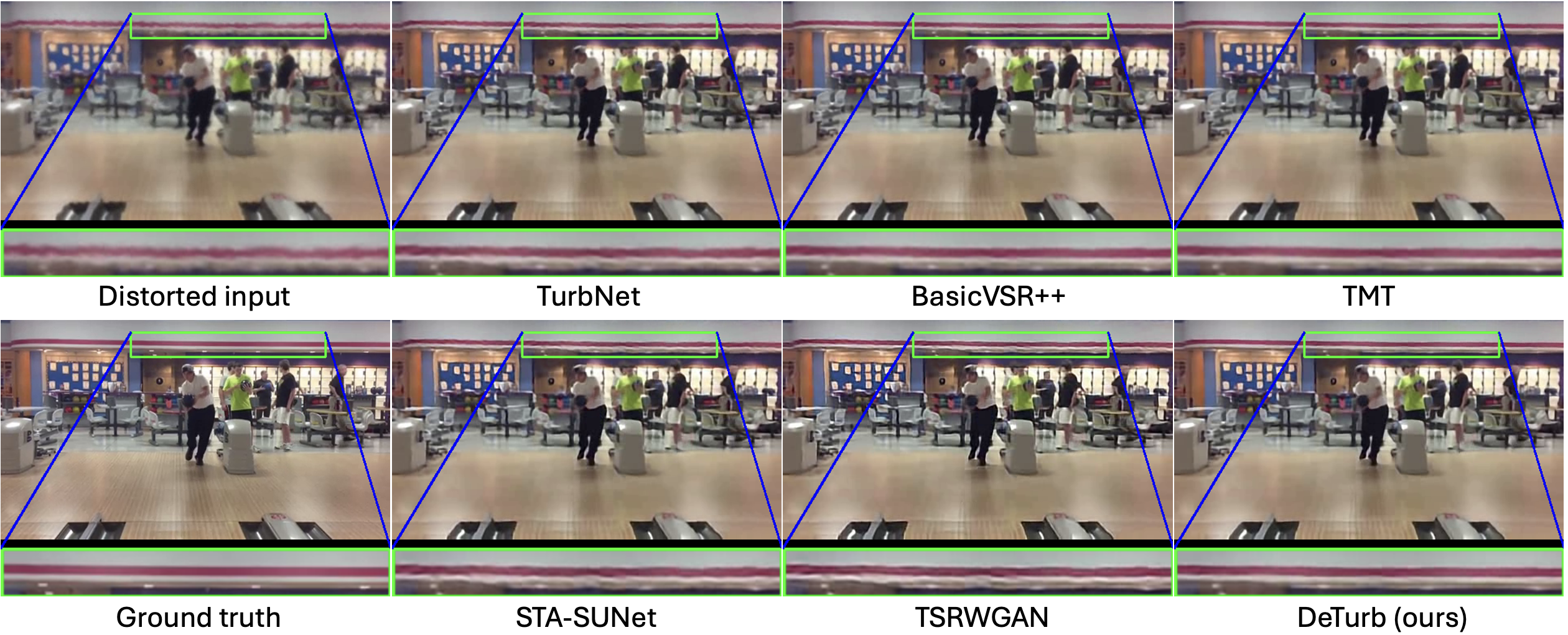}
	\caption{Subjective results of a synthetic scene. The bottom of each picture shows a magnified version of the straight lines.}
	\label{fig:syn_results}
\end{figure}

With ground truth available, we perform an objective assessment using PSNR, SSIM, and LPIPS measurements. Table~\ref{tab:static_dynamic} presents the average results, calculated by first averaging the results of all frames within each scene, and then averaging these scene results. This approach ensures that the results are not biased toward longer videos. The results clearly show that our method outperforms other existing methods for all metrics, highlighting that the deformable property facilitates a more flexible convolution operation, effectively modeling and correcting geometric distortions caused by atmospheric variations. Additionally, 3D Swin transformers demonstrate their effectiveness in deblurring and visual quality enhancement. Interestingly, BasicVSR++ delivers good results in PSNR and SSIM, but not in LPIPS. BasicVSR++ also exploits deformable convolutions, but applies optical flows for large motion beforehand. This should benefit dynamic scenes, which we plan to investigate further in future work. Example subjective results of the synthetic videos are shown in Fig. \ref{fig:syn_results}, where our visual results are sharper, and the zoomed-in areas of the straight lines reveal cleaner and clearer restored lines than other methods. This comparison obviously aligns with the objective results.

\begin{table}
    \centering
    \caption{Performance comparison on {static} and {dynamic} scenes using a synthetic dataset. Bold and underline indicate the best and second-best results, respectively.}
    \resizebox{\columnwidth}{!}{\begin{tabular}{l|ccc|ccc}
        \hline
        \multirow{2}{*}{Methods} & \multicolumn{3}{c|}{{Static Scenes}} & \multicolumn{3}{c}{{Dynamic Scenes}} \\
        \cmidrule(lr){2-4} \cmidrule(lr){5-7}
         & PSNR $\uparrow$ & SSIM $\uparrow$ & LPIPS $\downarrow$ & PSNR $\uparrow$ & SSIM $\uparrow$ & LPIPS $\downarrow$ \\
        \hline
        AT-Net~(\cite{Yasarla:ATNet:2021}) & 20.86 & 0.603 & 0.518 & 20.21 & 0.587 & 0.488 \\
        TurbNet~(\cite{10.1007/978-3-031-19800-7_25}) & 21.40 & 0.637 & 0.431 & 21.35 & 0.635 & 0.433 \\
        STA-SUNet~(\cite{Lin:Spatio:2024}) & 23.58 & 0.712 & 0.352 & 23.45 & 0.706 & 0.348 \\
        BasicVSR++~(\cite{Chan_2022_CVPR}) & {26.10} & 0.792 & 0.279 & \underline{26.14} & {0.790} & 0.280 \\
        TSRWGAN~(\cite{jin_chen_lu_chen_wang_liu_guo_bai_2021}) & 24.99 & 0.763 & 0.249 & 25.05 & 0.760 & {0.251} \\
        VRT~(\cite{Liang:VRT:2024}) & 25.85 & 0.782 & 0.218 & 25.78 & 0.766 & 0.266 \\
        TMT~(\cite{Zhang_TMT}) & 25.94 & \underline{0.795} & {0.202} & 26.09 & 0.767 & 0.264 \\
        DATUM~(\cite{zhang2024spatio}) & \underline{27.00} & 0.787 & \underline{0.198} & 27.35 & \textbf{0.819} & \underline{0.250}\\
        \hline
        DeTurb (ours) & \textbf{27.17} & \textbf{0.827} & \textbf{0.170} & \textbf{27.44} & \underline{0.815} & \textbf{0.242} \\
        \hline
    \end{tabular}}
    \label{tab:static_dynamic}
\end{table}

\subsection{Performance on real dataset}

For real atmospheric turbulence, we evaluated our proposed DeTurb and compared it with existing methods using both reference and no-reference metrics. The static scenes of the OTIS dataset come with ground truth, while the scenes of the CLEAR dataset come with pseudo ground truth; hence, objective assessment is possible. We employed NIQE scores~(\cite{6353522}) as a no-reference evaluation. NIQE is based on  a quality-aware set of statistical features derived from a straightforward yet effective natural scene statistic (NSS) model in the spatial domain. These features are extracted from a collection of natural, undistorted images. Lower scores indicate better perceptual quality.

\begin{figure}[tb]
\centering
 \includegraphics[width=\linewidth]{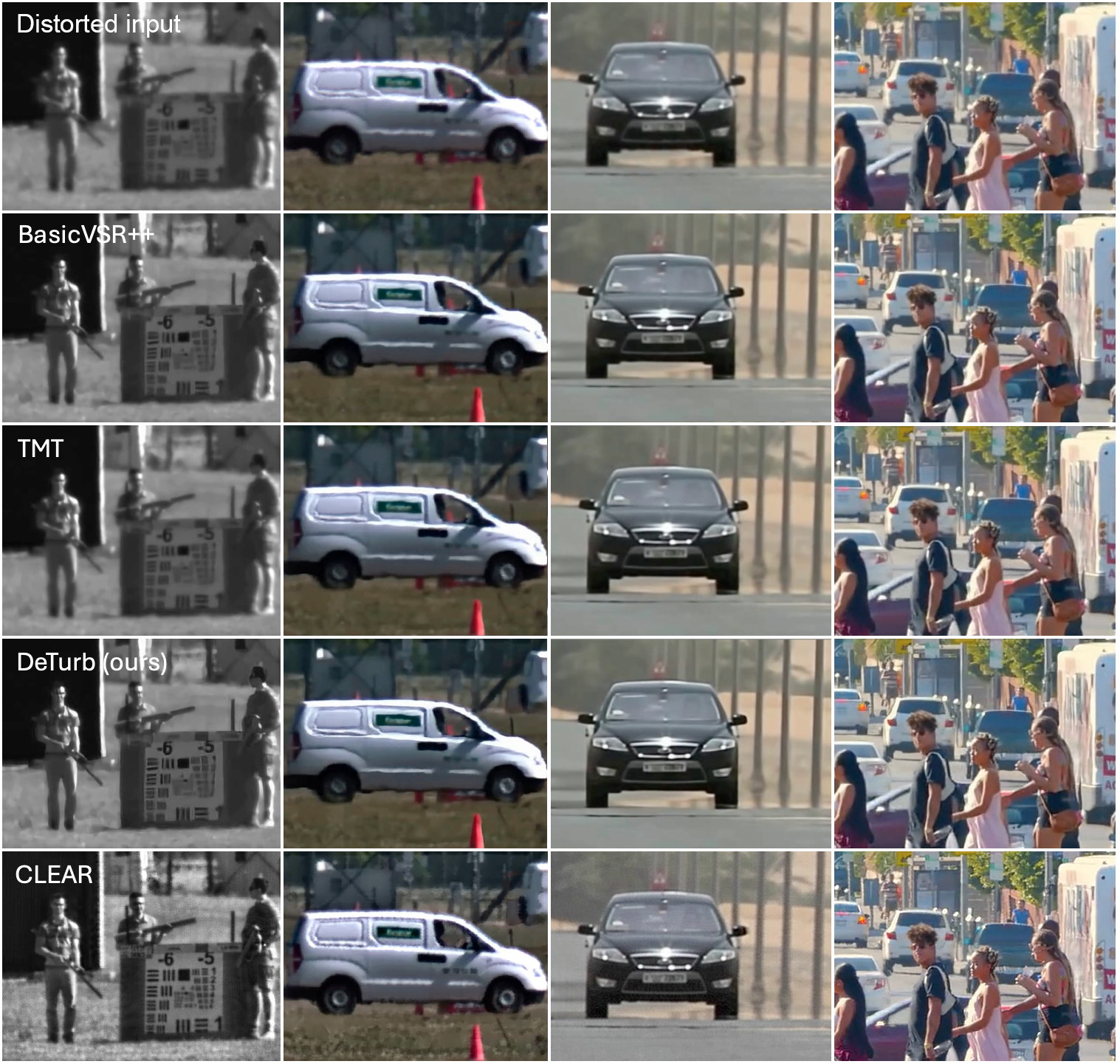}
	\caption{Subjective results of a real distorted scene. The left column shows a static scene, while the other columns depict dynamic scenes. From top to bottom, the rows display the distorted input, BasicVSR++ results, TMT results, our DeTurb results, and CLEAR results.}
	\label{fig:real_results}
\end{figure}

\begin{figure}
  \centering
  \includegraphics[width=0.5\linewidth]{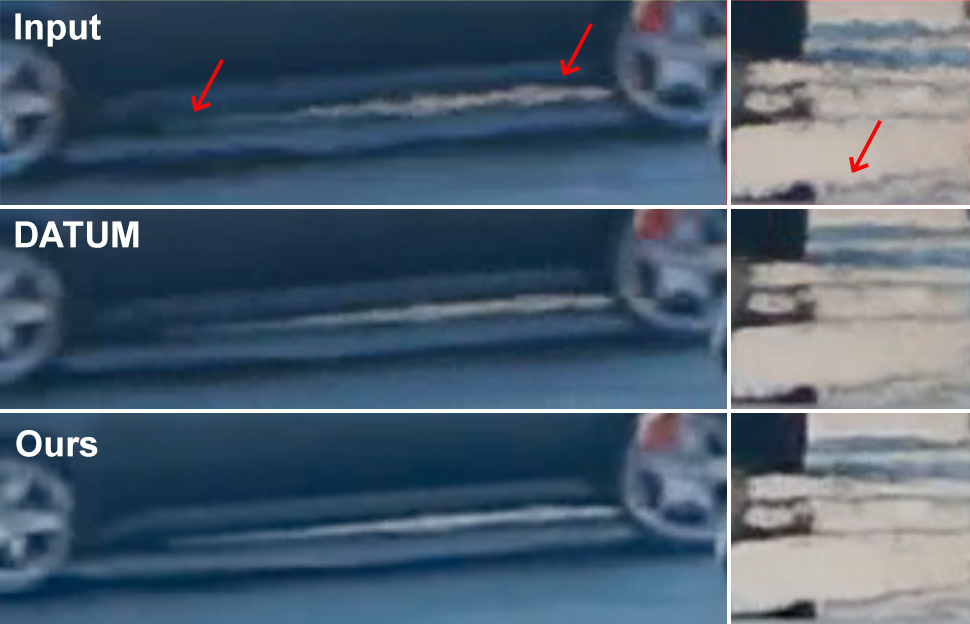}
   \caption{Subjective results (similar to Fig. 11 in the DATUM paper).}
   \label{fig:CompareWithDATUM}
\end{figure}

Table~\ref{tab:real_dynamic} shows the average results of these scenes. Comparing the two existing methods that performed best on the synthetic data, BasicVSR++ and TMT, our method achieves better scores across all metrics, particularly in LPIPS. For the no-reference metric NIQE, our method achieves approximately 9\% and 16\% higher scores, respectively. Fig.~\ref{fig:real_results} and Fig.~\ref{fig:CompareWithDATUM} illustrate examples of the restored scenes, with the last column showing the pseudo ground truth from CLEAR. Our method produces results with clearer edges and more readable text compared to BasicVSR++ and TMT. Although it cannot restore textures or text as well as CLEAR, our method achieves smoother edges than CLEAR. Overall, DeTurb outperforms other learning-based methods.

We found that when the atmospheric turbulence is low, DeTurb produces very sharp edges, achieving straight lines and clear curves, particularly at high contrast, such as text on road signs. However, strong distortions are more difficult to recover, as demonstrated in the last row where light rays propagate through the medium from a distance. Although the results appear sharper, random geometric distortion is still present in the results of all methods.

\begin{table}
    \centering
    \caption{Performance comparison on real atmospheric turbulence scenes. Bold and underline indicate the best and second-best results, respectively.}
    \resizebox{\columnwidth}{!}{\begin{tabular}{l|cccc|cccc}
        \hline
        \multirow{2}{*}{Methods} & \multicolumn{4}{c|}{{Static Scenes}} & \multicolumn{4}{c}{{Dynamic Scenes}} \\
        \cmidrule(lr){2-5} \cmidrule(lr){6-9}
         & PSNR $\uparrow$ & SSIM $\uparrow$ & LPIPS $\downarrow$ & NIQE $\downarrow$ & PSNR $\uparrow$ & SSIM $\uparrow$ & LPIPS $\downarrow$ & NIQE $\downarrow$ \\
        \hline
        Distorted inputs & - & - & - & 26.48 & - & - & - & 28.07 \\
        AT-Net~(\cite{Yasarla:ATNet:2021}) & 15.65 & 0.593 & 0.483 & 16.37 & 18.04 & 0.773 & 0.472 & 27.56 \\
        TurbNet~(\cite{10.1007/978-3-031-19800-7_25}) & 15.08 & 0.684 & 0.453 & 26.11 & 19.38 & 0.736 & 0.411 & 27.23 \\
        STA-SUNet~(\cite{Lin:Spatio:2024}) & 20.74 & 0.698 & 0.433 & 26.94 & 22.66 & 0.784 & 0.358 & 26.68\\
        BasicVSR++~(\cite{Chan_2022_CVPR}) & {21.66} & {0.754} & 0.232 & {26.72} & \underline{26.84} & \underline{0.845} & 0.206 & {25.03} \\
        TSRWGAN~(\cite{jin_chen_lu_chen_wang_liu_guo_bai_2021}) & 16.87 & 0.710 & {0.228} & {26.72} & 24.73 & 0.809 & 0.130 & 24.88\\
        VRT~(\cite{Liang:VRT:2024}) & 15.25 & 0.620 & 0.338 & 28.46 & 25.26 & 0.818 & 0.133 & 25.63 \\
        TMT~(\cite{Zhang_TMT}) & 15.17 & 0.611 & 0.329 & 29.21 & 25.59 & 0.822 & {0.122} & 27.98  \\
        DATUM~(\cite{zhang2024spatio}) & \textbf{24.65} & \underline{0.827} & \underline{0.204} & \underline{25.01} & 26.46 & 0.838 & \underline{0.096} & \textbf{22.72} \\
        \hline
        DeTurb (ours) & \underline{24.53} & \textbf{0.841} & \textbf{0.128} & \textbf{24.34} & \textbf{26.99} & \textbf{0.847} & \textbf{0.088} & \underline{22.77} \\
        \hline
    \end{tabular}}
    \label{tab:real_dynamic}
\end{table}

\subsection{Ablation study}

The results of the ablation study are shown in Table~\ref{tab:ablation}. First, we investigated the influence of deformable 3D convolutions by replacing them with depth-wise 3D convolutions, as used in TMT (\cite{Zhang_TMT}). All metrics indicated a performance reduction, particularly for LPIPS. Next, we tested the impact of the 3D Swin transformation by replacing it with 2D Swin transformers (\cite{Liu_2021_ICCV}). Although the overall performance was reduced, it was not as significant as the decrease observed when replacing deformable 3D convolutions with depth-wise 3D convolutions. This highlights the importance of addressing spatiotemporal distortions caused by atmospheric turbulence. This significance is further confirmed by the substantial reduction in restoration quality when the non-rigid registration module is removed, which has a greater impact than removing the feature fusion module from the pipeline.

\begin{table}[tb]
    \centering
    \caption{Ablation study showing the effects of replacing or removing certain modules in the proposed framework. The results are averaged from both synthetic and real scenes, except for NIQE, which is computed only on the real scenes.}
    \resizebox{\columnwidth}{!}{\begin{tabular}{l|cccc|cccc}
        \toprule
        \multirow{2}{*}{Methods} & \multicolumn{4}{c|}{{Static Scenes}} & \multicolumn{4}{c}{{Dynamic Scenes}} \\
        \cmidrule(lr){2-5} \cmidrule(lr){6-9}
         & PSNR $\uparrow$ & SSIM $\uparrow$ & LPIPS $\downarrow$ & NIQE $\downarrow$ & PSNR $\uparrow$ & SSIM $\uparrow$ & LPIPS $\downarrow$ & NIQE $\downarrow$ \\
        \hline
        Deform3D $\rightarrow$ DW Conv 3D & 24.25 & 0.772 & 0.249 & 25.73 & 27.05 & 0.789 & 0.231 & 25.08 \\
        3D Swin $\rightarrow$ 2D Swin & 25.15 & 0.800 & 0.188 & 25.75 & 27.13 & 0.816 & 0.193 & 23.72 \\
        wo Non-rigid & 23.88 & 0.761 & 0.261 & 26.02 & 26.87 & 0.795 & 0.246 & 25.12 \\
        wo Fusion & 24.07 & 0.778 & 0.258 & 25.70 & 26.53 & 0.800 & 0.228 & 24.89 \\
        \hline
        DeTurb & 25.85 & 0.834 & 0.149 & 24.34 & 27.22 & 0.831 & 0.165 & 22.77 \\
        \bottomrule
    \end{tabular}}
    \label{tab:ablation}
\end{table}

\begin{figure}[tb]
\centering
	\includegraphics[width=0.9\linewidth]{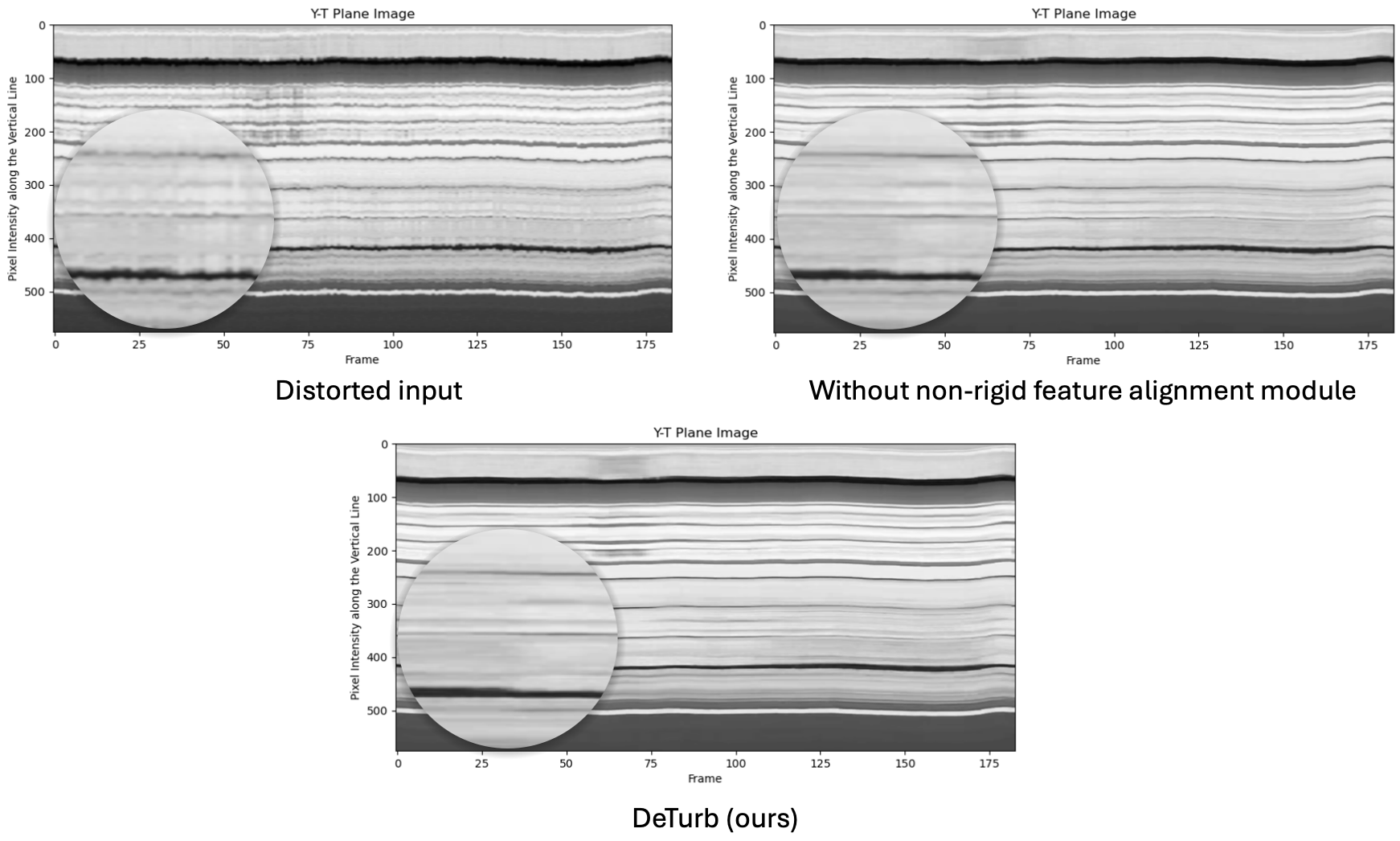}
	\caption{Example $y$-$t$ planes of static `Man' scene restored without and with the non-rigid registration module}
	\label{fig:yt_planes}
\end{figure}

For clearer visualization, Fig. \ref{fig:yt_planes} shows the $y$-$t$ plane images, illustrating how a specific line or point at a particular $x$ position within an image evolves over time across video frames. This provides a view of the stabilization achieved by turbulence mitigation models. We selected the middle point of the width. The left image is from the original distorted scene, showing clear vertical streaks and inconsistencies. The feature fusion model alone shows improvement, with noticeably reduced streaking and clearer continuity of lines, but some random geometric distortions are still present, as seen in the middle figure. The entire pipeline, including the non-rigid registration, exhibits the highest level of clarity and consistency, effectively mirroring the original structure of the scene with superior stabilization of vertical elements.

\subsection{Computing Budget}

We include an assessment of the inference-time computing budget for each model. This analysis is conducted on a single NVIDIA 4090 GPU, providing a standardized basis for comparing the computational demands of each model. Table~\ref{tab: inference-time} summarizes the computing resources and time required for each model. Although none of the learning-based methods currently meet real-time requirements, they are significantly faster than conventional model-based methods such as SID and CLEAR.

\begin{table}[tb]
	\begin{center}
	\caption{Inference time computing budget, measured per frame on a 256$\times$256 resolution image (average speed calculated from five test trials).} 
	\begin{tabular}{c|ccc}
		\toprule
		Methods	& $\#$ parameters (M)  & FLOPs/frame (G) & speed (s) \\
		\hline
        SID (\cite{6178259}) & - & - & 132.32 \\
        CLEAR~(\cite{ANANTRASIRICHAI202369}) & - & - & 25.37 \\ \hline
		TurbNet~(\cite{10.1007/978-3-031-19800-7_25}) & 26.60 & 190.69  & 5.79 \\
		STA-SUNet~(\cite{Lin:Spatio:2024}) & 21.82 & -& 2.04\\
		BasicVSR++~(\cite{Chan_2022_CVPR}) & 9.76 & 127.30 & 1.20 \\	TSRWGAN~(\cite{jin_chen_lu_chen_wang_liu_guo_bai_2021}) & 46.28 & 2,836 & 2.58 \\
		VRT~(\cite{Liang:VRT:2024}) & 18.32 & 7,759 & 7.35 \\
		TMT~(\cite{Zhang_TMT}) & 23.92 & 1,304 & 2.37 \\
		\hline
		DeTurb (ours) & {58.79}  & {1,975} & {2.55}\\
		\bottomrule
	\end{tabular}
	\label{tab: inference-time}
 \end{center}
\end{table}

\section{Conclusion}

This paper introduces DeTurb, a novel framework for atmospheric turbulence reduction. The proposed framework consists of two modules: non-rigid registration and feature fusion. The first module uses deformable 3D convolutions to estimate flow and mitigate random geometric distortion across several distorted frames. The second module, utilizing a UNet-like architecture of 3D Swin transformers, further sharpens and enhances the details of the current frame. Experimental results demonstrate that DeTurb outperforms existing methods specifically designed for atmospheric turbulence problems as well as methods proposed for video restoration and enhancement. However, there is still room for improvement in situations of strong atmospheric turbulence.

\bibliographystyle{plainnat}
\bibliography{template}

\end{document}